\def\year{2018}\relax
\documentclass[letterpaper]{article} 
\pdfoutput=1
\usepackage{aaai18}  
\usepackage{times}  
\usepackage{helvet}  
\usepackage{courier}  
\usepackage{url}  
\usepackage{graphicx}  
\usepackage{booktabs}       
\begin{document}
%

\title{DeepHeart: Semi-Supervised Sequence Learning for Cardiovascular Risk Prediction}
\author{\normalsize{Brandon Ballinger, Johnson Hsieh, Avesh
Singh}, \\ \normalsize{\textbf{Nimit Sohoni, Jack Wang}} \\
Cardiogram\\
San Francisco, CA \\
\And
\normalsize{Geoffrey H. Tison, Gregory M. Marcus,} \\
\normalsize{\textbf{Jose M. Sanchez, Carol Maguire}} \\
\normalsize{\textbf{Jeffrey E. Olgin, Mark J. Pletcher}} \\
Department of Medicine \\
University of California \\
San Francisco, CA}
\maketitle
\begin{abstract}
  We train and validate a semi-supervised, multi-task LSTM on 57,675 person-weeks of data from off-the-shelf
  wearable heart rate sensors, showing high accuracy at detecting multiple medical conditions,
  including diabetes (0.8451), high cholesterol (0.7441), high blood pressure (0.8086), and sleep apnea (0.8298).
  We compare two semi-supervised training methods, semi-supervised sequence learning and heuristic pretraining,
  and show they outperform hand-engineered biomarkers from the medical literature. We believe
  our work suggests a new approach to patient risk stratification based on cardiovascular risk scores
  derived from popular wearables such as Fitbit, Apple Watch, or Android Wear.
\end{abstract}
\section{Introduction}

In medicine, each label represents a human life at risk ---for example, a person who recently suffered
a heart attack or experienced an abnormal heart rhythm.  As a result, even widely-used predictive
models like CHA2DS2-VASc may be derived from as few as 25 positive labels \cite{lib10}. However, popular
wearables, such as Fitbit and Apple Watch, generate trillions of unlabeled sensor data points
per year, including rich signals like resting heart rate and heart rate variability, which have
been shown to correlate with health conditions as diverse as diabetes, sleep apnea, atrial fibrillation,
heart failure, sudden cardiac death, and irritable bowel syndrome \cite{kamath2012heart}.

Using consumer-grade heart rate sensors in a medical context presents several challenges. First,
the sensors themselves have significant error \cite{gillinov16}. Second, they vary the rate of
measurement to preserve battery life. Third, since wearables are used in an
ambulatory setting, daily activities like walking, exercise, stress, consuming alcohol, or drinking
coffee (Figure 1) may confuse simple heuristics.

Deep neural networks \cite{lecun2015deep} have shown high accuracy at pattern recognition from noisy, complex
inputs, including automated detection of diabetic retinopathy from images \cite{gulshan2016development}, skin cancer from
mobile phone cameras \cite{esteva2017dermatologist}, and the onset of health conditions from electronic medical records
\cite{razavian2016multi,lipton2015learning,choi2016doctor,che2015deep}. However, purely supervised deep learning requires many
labeled examples. No such database exists for wearable data, and in most hospital systems,
recruiting even a hundred patients into a new medical study is labor-intensive and expensive.

To solve this, we take a two-pronged approach. First, we employ a mobile phone application to
recruit 14,011 participants from across the world, collecting 200 million unlabeled sensor
measurements. Data from these participants is then used in one of two semi-supervised training
procedures. In semi-supervised sequence learning \cite{dai2015semi}, an LSTM \cite{hochreiter1997long} is pretrained
as a sequence autoencoder; weights from pretraining are used as initialization for a second
supervised phase, making use of a limited pool of labeled data from participants with known
diagnoses. In heuristic pretraining, the neural network is instead pretrained to compute
heart-rate-derived biomarkers from the medical literature, partially bridging the gap between
feature engineering and deep learning.

The rest of this paper is organized as follows. We first describe a novel data set derived
from 14,011 participants with wearable heart rate monitors, recruited in partnership with the Health
eHeart Study of the University of California, San Francisco. We then describe DeepHeart, our model
architecture, and the semi-supervised training methods used,
semi-supervised sequence learning and heuristic pretraining. Then we present experimental
results showing that semi-supervised deep neural networks achieve higher accuracy in detecting
diabetes, sleep apnea, high blood pressure, and high cholesterol than several strong baselines
derived from biomarkers in the medical literature. We conclude by describing some of the
historical barriers to wide-scale deployment of artificial intelligence in medicine and how we plan
to address them in future work.

\section{Data and Study Cohort}

Electronic medical records contain important discrete events, such as lab tests and diagnoses,
while continuous wearable data from mobile health platforms, such as Apple HealthKit or Google Fit,
cover the majority of the patient's time outside of a medical setting. In order to train a deep
neural network to predict diagnoses given wearable data, we sought to combine the two
data sources.

\begin{figure*}[ht]
  \begin{center}
    \includegraphics[width=400pt]{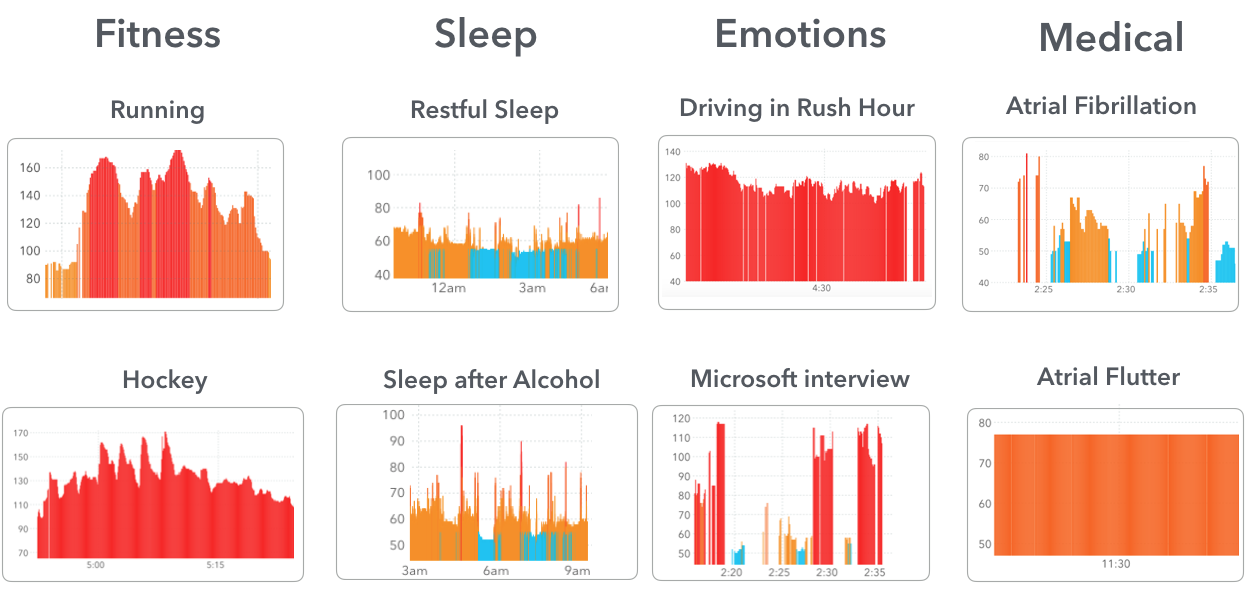}
  \end{center}
  \caption{Examples of wearable heart rate data showing different physiological states: exercise,
           sleeping, stress, and medical conditions.}
  \label{fig:examples_of_heart_rate_data}
\end{figure*}

\subsection{Study Cohort}

We recruited 14,011 users of a popular Apple Watch app into an online, IRB-approved study run
in partnership with the cardiology department of the University of California, San Francisco \cite{tison2017B}. Participants completed a medical history,
including previous diagnoses, blood test results, and medications. A mobile app---Cardiogram---integrated with
HealthKit continuously stored, processed, and displayed participant heart rate, step count, and other activity
data.

We chose to focus on four highly prevalent but often undiagnosed conditions associated with cardiovascular risk:
high cholesterol, hypertension (high blood pressure), sleep apnea, and diabetes. 45.8\% of diabetes
cases are undiagnosed \cite{beagley2014global}. Sleep apnea is estimated to affect 22 million people in the US
alone, over 80\% of whom are undiagnosed \cite{young1997estimation}. Together,
high blood pressure and high cholesterol account for 17.3\% of global deaths \cite{alwan2011global}, as they are both major
risk factors for heart attacks.

\begin{table}[hb]
  \caption{Cohort Statistics}
  \label{cohort-stats}
  \centering
  \begin{tabular}{llll}
    \toprule
    Condition            & Number of  & Person-    & Person-   \\
                         & People     & Weeks of   & Weeks of  \\
                         &            & Training   & Testing   \\
                         &            & Data       & Data      \\
    \cmidrule{1-4}
    Unlabeled            & 1,462      & 10,616     & N/A    \\
    \cmidrule{1-4}
    High Cholesterol     & 2,331      & 9,220      & 3,243  \\
    Hypertension         & 2,230      & 7,651      & 2,762  \\
    Sleep Apnea          & 1,016      & 3,662      & 1,141  \\
    Diabetes             &  462       & 1,678      &  644   \\
    \bottomrule
  \end{tabular}
\end{table}

\subsection{Multi-Channel, Multi-Timescale Sensor Data}

Consumer wearables produce multiple channels of data---most commonly, step count measured using
a wrist-worn accelerometer and optical heart rate measured using photoplethysmography. For example,
the Apple Watch measures heart rate every five seconds when in workout mode and roughly every five
minutes during other times of the day. Likewise, step counts are recorded every few minutes
while the user is walking.

We encoded sensor measurements into a tensor $X_{u,t,c}$ with indices for the user-week $u$,
timestep $t$, and input channel $c$. Heart rate and step count were encoded as separate input
channels. To accommodate variable timescales, we encoded intra-channel time-deltas as a special
dt channel:

  \begin{equation} dt_{transformed} = 0.1 log(\frac{dt}{5000}) \end{equation}

Without transforming dt, the model does not train due to the wide spread of values (ranging from 5000 to 28,800,000).

Sequences contained up to 4096 timesteps of sensor data, and shorter sequences were zero-padded.

\subsection{Pre-Processing}
Each participant was randomly assigned to either the training, tuning, or testing set. Each
participant's data was then split into week-long chunks, and any weeks with $\leq 672$ heart rate
measurements ($\leq 8$ hours per day of background measurements) or $\leq 30$ minutes of continuous
heart rate recordings were omitted. After filtering, there were 57,675 person-weeks of data in total,
divided into 33,628 for training, 18,555 for tuning, and 12,790 for validation. The training, tuning,
and testing sets contained completely disjoint sets of participants.

\subsection{Multi-Task Labeled Output}

Similar to X, our output is a tensor $Y_{u,t,c}$ with indices for the user-week $u$,
timestep $t$, and output channel $c$. There are separate output channels for prevalent cases of each
condition: high cholesterol, hypertension, sleep apnea, and diabetes. Diagnoses are derived from validated health surveys collected
in partnership with the Health eHeart Study of the University of California,
 San Francisco. In addition, we included an output channel of ECG readings\footnote{Our research builds on previous work to detect atrial fibrillation using ECGs. We chose to keep
this output task so our model could benefit from ECG readings.}. ECG readings were aligned to the
closest sensor measurement in X, and diagnoses were aligned to the last timestep in X.
Each $Y_{t}[task]$ is a discrete label: either positive (user has condition, or abnormal ECG reading)
or negative (user is not diagnosed with condition, or ECG reading is normal).
We used output masking to score the models only when $Y_{t}[task]$ has a value.

\section{Model Architecture}

Figure 2 shows our model architecture. The inputs are multi-channel, multi-timescale sensor measurements,
and the outputs are multi-task, multi-timescale diagnoses. Table 3 summarizes hyperparameter
tuning experiments.

\begin{figure}
  \begin{center}
    \includegraphics[width=239pt]{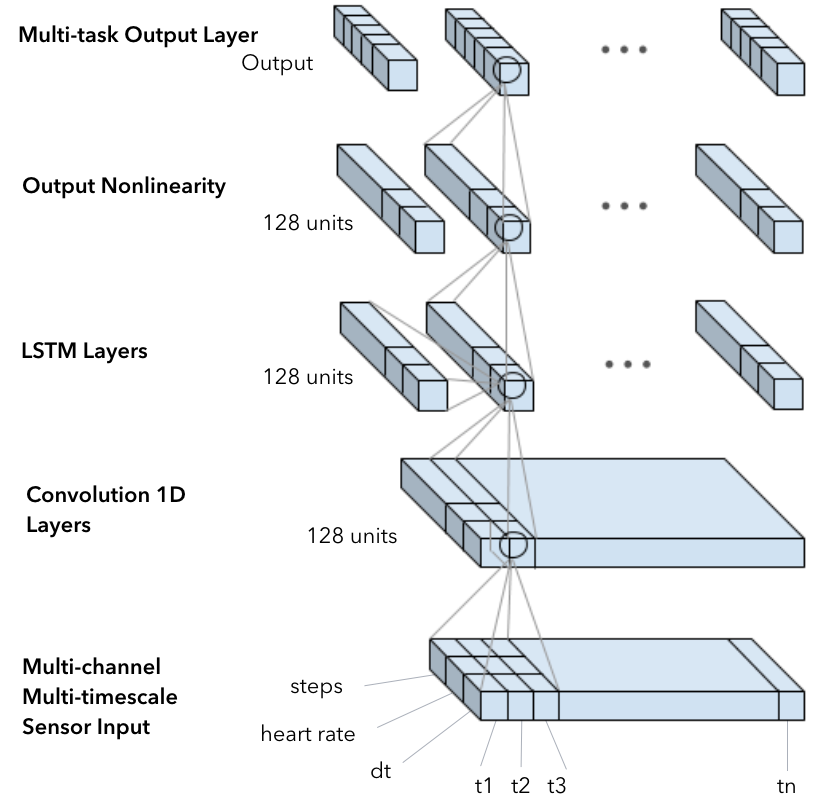}
  \end{center}
  \caption{DeepHeart architecture.}
  \label{fig:architecture}
\end{figure}

\subsection{Temporal Convolutions}

The first three layers are temporal convolutions, which have been shown to efficiently extract
features and model temporal translation invariance \cite{amodei2016end}. The first layer has a wide filter
length of 12, similar to AlexNet \cite{krizhevsky2012imagenet}, and the next two layers use the residual units
\cite{he2016identity} with a filter length of 5. Each convolutional layer contains 128 convolutional channels.
After each convolutional layer, we apply dropout with probability 0.2 to prevent overfitting, and
apply max pooling with pool length 2 to reduce dimensionality.

\subsection{Bidirectional LSTM}

The convolutional layers are followed by four bidirectional LSTM layers that
model longer temporal patterns in wearable sensor data and their corresponding multi-timescale diagnoses.
Each bidirectional LSTM layer contains 128 units (64 in each direction).

A dropout of 0.2 is applied to this final LSTM layer, and the result is run through a simple output
unit: a convolution of filter length 1. The purpose of this convolution is to map the outputs of the LSTM
into a single prediction per task, per timestep. For this final layer, we use a tanh activation so outputs range from -1 to 1.

\section{Training Methods}

In addition to ordinary supervised training, we experimented with two pretraining
architectures. Semi-supervised training is advantageous here because unlabeled data from wearables is abundant, but
medical condition labels are more scarce.

\subsection{Unsupervised Sequence Pretraining}

Semi-supervised sequence learning has been shown to improve performance when presented with large
amounts of unlabeled training data \cite{dai2015semi}. We experimented with this approach by first
training a sequence-to-sequence autoencoder, consisting of the 3 convolutional and 4 recurrent
layers. We then use the weights from the autoencoder as initializing parameters for the
corresponding layers in the supervised architecture. In order to force the autoencoder to
not merely memorize the input, we added Gaussian noise to the input sequence.

\subsection{Weakly-Supervised Heuristic Pretraining}

Previous work has applied statistical methods on beat-to-beat variability to detect heart
arrhythmias \cite{mcmanus2013novel}. Inspired by this, we synthesized a dataset for pretraining using
a time-windowed heart rate variability metric. We defined four output tasks, with window
sizes of 5 seconds, 30 seconds, 5 minutes, and 30 minutes, computed the average absolute
difference between successive heart rate measurements, and used these as output channels in
a weakly-supervised pretraining phase. The resulting weights were used for initialization
in supervised training. In this way, we use pretraining as a technique to bridge feature
engineering and deep learning.

\section{Experiments}

To evaluate our architecture, we ran a series of experiments summarized in Table \ref{table:results}.
For each condition on our data set, we report the c-statistic---the area under the ROC curve---on
the test set. Each experiment used the Adam optimizer \cite{kingma2014adam} and a squared error loss, since we
found squared error to be more numerically stable than cross-entropy.

\begin{table*}[ht]
  \caption{Experiment Results (c-statistics)}
  \label{table:results}
  \centering
  \begin{tabular}{llllll}
    \toprule
    Name                          &   Diabetes      & High Cholesterol & Sleep Apnea     & High Blood Pressure \\
    \midrule
    Logistic Regression           &   0.7906        & 0.5941             & 0.6583        & 0.6389              \\
    Support Vector Machine        &   0.5722        & 0.5996             & 0.5490        & 0.6106              \\
    Decision Tree                 &   0.4142        & 0.5152             & 0.6415        & 0.6625              \\
    Random Forest                 &   0.5515        & 0.5647             & 0.5907        & 0.6310              \\
    Multi-layer Perceptron        &   0.7846        & 0.4327             & 0.6172        & 0.7195              \\
    \midrule
    LSTM, No Pretraining          &   \textbf{0.8451}  & 0.6736           & 0.8041          & 0.7991              \\
    \midrule
    LSTM, Heuristic Pretraining     & 0.8366        & 0.7148           & 0.7951           & 0.7427              \\
    LSTM, Unsupervised Pretraining  & 0.7998        & \textbf{0.7441}  & \textbf{0.8298}  & \textbf{0.8086}     \\
    \bottomrule
  \end{tabular}
\end{table*}

\subsection{Baselines}

We compared our architecture with multiple machine learning algorithms using hand-engineered
biomarkers derived from the medical literature, including resting heart rate \cite{fox2007resting},
time-windowed average heart rate \cite{kamath2012heart}, time-windowed standard deviation of heart rates
\cite{kamath2012heart}, time-windowed spectrum entropy \cite{kamath2012heart}, time-windowed root mean squared of successive
beats per minute differences \cite{mcmanus2013novel}, and time-windowed entropy of BPM differences \cite{mcmanus2013novel}. We used 5-minute
and 30-minute time windows, and included global versions of standard deviation and RMS of successive
BPM differences, for 13 features in total. We trained several standard machine learning algorithms
on all of these features, including scikit-learn's \cite{pedregosa2011scikit} implementation of logistic
regression, support vector machines, decision trees, random forests, and multi-layer perceptrons.

\subsection{LSTM without Pretraining}

We first trained a purely-supervised LSTM without pretraining, which performed significantly better than
the baselines on sleep apnea (0.80 vs 0.55-0.66), hypertension (0.80 vs 0.61-0.72), high cholesterol
(0.67 vs 0.43-0.60), and diabetes (0.85 vs 0.41-0.79). The gap is smallest for diabetes, with the least
amount of labeled data, suggesting potential benefit from combining hand-engineered features with
deep learning. Previous work from the medical community has suggested that diabetes may impact heart
rate variability through effects on the autonomic nervous system \cite{kudat2006heart}. Our work builds on top of these previous findings, and suggests that
heart rate variability changes driven by diabetes can be detected via consumer, off-the-self wearable heart rate sensors.

\subsection{Heuristic Pretrained LSTM}

When the supervised LSTM is initialized with weights from heuristic pretraining, we saw noticeable
improvements for high cholesterol (0.71 vs 0.67), neutral results for sleep apnea and diabetes,
and a significant loss for high blood pressure (0.74 vs 0.80). The gain on high cholesterol is particularly
surprising, given that the baseline
results, which are based on similar hand-engineered heuristic features, did not perform well.
One explanation is that the neural network is able to find non-linear relationships for the underlying
patterns that generate these heuristic features, and the insights discovered by the heuristic
pretraining were useful for predicting high cholesterol.

On high blood pressure, the heuristic-pretrained LSTM performs significantly worse than the
LSTM without pretraining. One explanation is that the heuristic of average BPM differences is
a poor predictor of high blood pressure, and gradient descent cannot escape from this space of
poorly-performing predictors.

\subsection{Semi-supervised Sequence Learning}

The last experiment applied semi-supervised sequence learning: we first pretrained DeepHeart
as a sequence autoencoder, using the encoder weights as initialization for a second, supervised
phase, as described in \cite{dai2015semi}. This approach resulted in a significant improvement
on high cholesterol, high blood pressure, and sleep apnea.

The 0.83 c-statistic at detecting sleep apnea is surprising given that few participants sleep
with their watch. This suggests that wearable-derived biomarkers measured during the day capture
a distinct signature of sleep apnea.

Among all four health conditions, high cholesterol has the least direct physiological relationship
with heart rate variability in the medical literature. While we are able to detect high cholesterol
with c-statistic of 0.74, it is the lowest performing prediction out of the four disease
states, suggesting that much of the correlation may be a reflection of confounding variables such as
age, sex, or usage of medications like beta blockers. Nevertheless, a c-statistic of 0.74 from
consumer-grade wearables alone is a surprising and novel finding.

\subsection{Effect of Varying Amounts of Labeled Data}

To quantify the impact of each pretraining technique, we evaluated DeepHeart's performance
when trained on varying fractions of the labeled data (5\%, 10\%, 20\%, 50\%, 70\%, and 100\%), using no
pretraining, heuristic pretraining, and semi-supervised sequence learning (Figure \ref{fig:perf_vs_pct_labels}).

For hypertension and sleep apnea, unsupervised pretraining yields a 10x improvement in data-efficiency:
the DNN achieves nearly the same accuracy with 10\% of labeled data using unsupervised pretraining as it
does with 100\% of labeled data using no pretraining. For diabetes and cholesterol, the effect is muted.
For diabetes, this is due to a scarcity of labels (1,678, vs 3,662-9,220 for the other
conditions). For high cholesterol, the lack of correlation between the number of labels and
accuracy of the algorithm likely confirms that high cholesterol is primarily detected through the
effect of confounding variables such as age, sex, and medications.

\begin{figure}[h]
  \begin{center}
    \includegraphics[width=239pt]{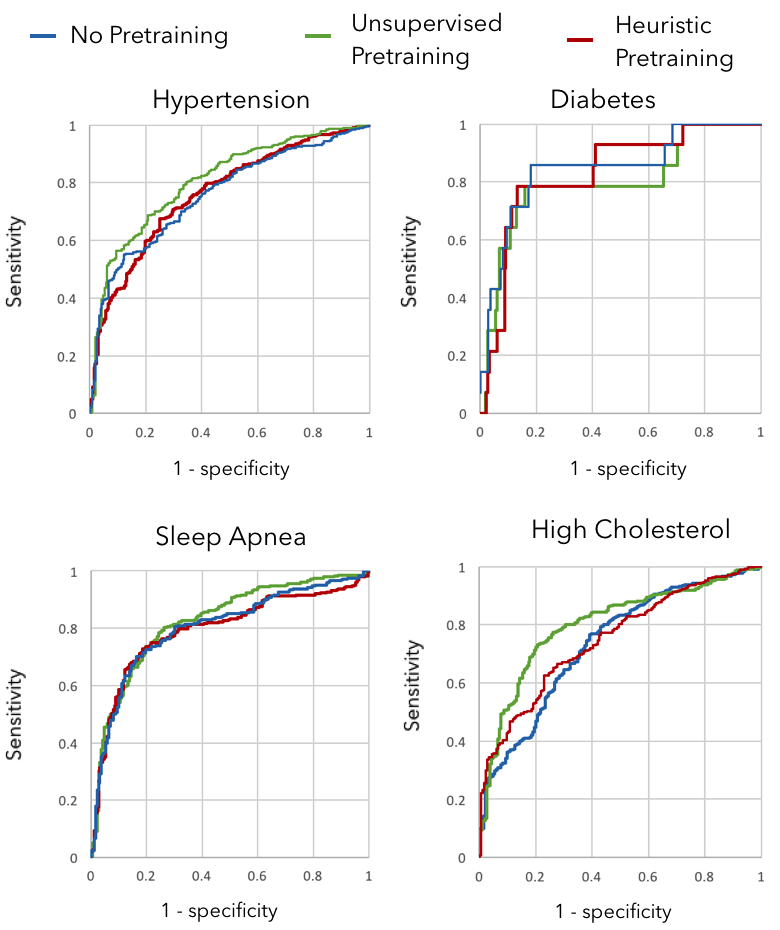}
  \end{center}
  \caption{ROC curves (sensitivity-specificity curves) comparing the accuracy of heuristic pretraining
  and semi-supervised pretraining to the strongest overall baseline in detecting diabetes, sleep apnea, high
  cholesterol, and high blood pressure.}
  \label{fig:roc_curves}
\end{figure}

The shape of the ROC curves (Figure \ref{fig:roc_curves}) are also interesting to note. It is clear that the
semi-supervised sequence LSTM performs very well for detecting diabetes, sleep apnea, and hypertension,
three of the most highly prevalent conditions with undiagnosis rates of 45.8\%, 80\%, and 20\% respectively.
For high cholesterol, although the c-statistic is lower, there are high-precision operating points
which rule out more than half of the study population. This suggests wearables could be used to
focus diagnostic testing on a high-risk subset of the population, leading to cost-effective screening for
metabolic syndrome and getting at-risk patients care sooner.

\begin{figure}[h]
  \begin{center}
    \includegraphics[width=239pt]{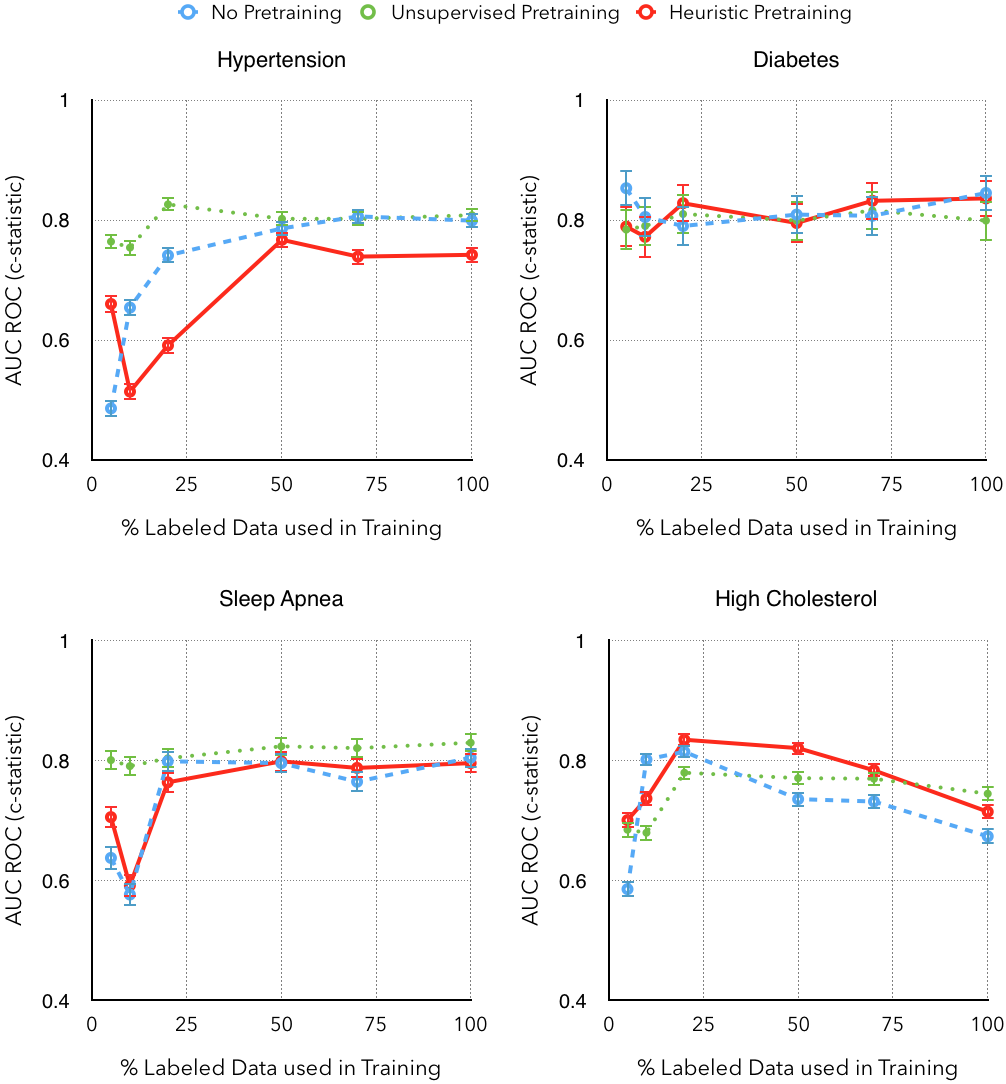}
  \end{center}
  \caption{Accuracy as a function of percentage of training labels used for each condition. Sleep apnea
  and hypertension, which show the greatest effect, have both a direct physiological connection to heart
  rate variability and a large amount of labeled data in our training set. For diabetes and high cholesterol,
  the effect of adding more labeled data is muted due to the low absolute number of labels and less direct
  physiological connection, respectively. Error bars show the 95th percentile confidence interval.}
  \label{fig:perf_vs_pct_labels}
\end{figure}

\subsection{Ablative Analysis of Input Channels}
To determine the impact of each input channel, we tested models using heart rate and
step count in isolation.

A model trained only on heart rate timeseries performed significantly worse
on diabetes and high blood pressure (decrease of 0.0653 and 0.0721 in AUC, respectively),
and the same on sleep apnea and high blood pressure (within the 95\% confidence
interval), suggesting that heart rate is being interpreted as a response to activity.
Within the medical literature, heart rate recovery after exercise is associated with lower risks
of cardiovascular disease \cite{morshedi2002heart}. A model using only aggregate step counts achieved an AUC
of 0.7011 (diabetes), 0.5811 (sleep apnea), 0.5714 (hbp), and 0.5601 (high cholesterol),
suggesting that although overall physical activity plays an important role in predicting
the onset of disease, the heart's response to physical activity is a more salient biomarker
that can be captured using deep learning.

\begin{table*}[b]

\begin{tabular}{llllllllll}
\hline
 Width & Conv Depth & LSTM Depth & Initial Filter &   AF &   Diabetes &   High BP &   Apnea &   Chol & Avg \\
\hline
  32 & 2  & 2  & 12   &                    0.7548 &     0.8709 & 0.8494 &        0.8610 &      0.8549  & 0.8382 \\
  32 & 2  & 2  & 5    &                    0.7340 &     0.8724 & 0.8375 &        0.8594 &      0.8541  & 0.8315 \\
  32 & 2  & 4  & 12   &                    0.7621 &     0.8748 & 0.8455 &        0.8664 &      0.8511  & 0.8400 \\
  32 & 2  & 4  & 5    &                    0.7633 &     0.8659 & 0.7995 &        0.8567 &      0.8257  & 0.8222 \\
  32 & 4  & 2  & 12   &                    0.7480 &     0.8540 & 0.8502 &        0.8583 &      0.8352  & 0.8291 \\
  32 & 4  & 2  & 5    &                    0.7652 &     0.8622 & 0.8258 &        0.8592 &      0.8211  & 0.8267 \\
  32 & 4  & 4  & 12   &                    0.7440 &     0.8711 & 0.8233 &        0.8605 &      0.8442  & 0.8286 \\
  32 & 4  & 4  & 5    &                    0.7336 &     0.8628 & 0.8403 &        0.8626 &      0.8444  & 0.8287 \\
  64 & 2  & 2  & 12   &                    0.7317 &     0.8696 & 0.8494 &        0.8599 &      0.8594  & 0.8340 \\
  64 & 2  & 2  & 5    &                    0.7514 &     0.8719 & 0.8544 &        0.8585 &      0.8525  & 0.8377 \\
  64 & 2  & 4  & 12   &                    0.7891 &     0.8647 & 0.8444 &        0.8597 &      0.8467  & 0.8409 \\
  64 & 4  & 2  & 12   &                    0.7903 &     0.8517 & 0.8550 &        0.8622 &      0.8513  & 0.8421 \\
  64 & 4  & 2  & 5    &                    0.7855 &     0.8655 & 0.8483 &        0.8559 &      0.8442  & 0.8399 \\
  64 & 4  & 4  & 12   &                    0.8079 &     0.8805 & 0.8453 &        0.8584 &      0.8647  & \textbf{0.8514} \\
  64 & 4  & 4  & 5    &                    0.7958 &     0.8756 & 0.8434 &        0.8632 &      0.8618  & 0.8480 \\
  128 & 2  & 2  & 5   &                    0.7770 &     0.8653 & 0.8529 &        0.8607 &      0.8549  & 0.8422 \\
  128 & 2  & 4  & 12  &                    0.8212 &     0.8718 & 0.8262 &        0.8604 &      0.8361  & 0.8431 \\
  128 & 2  & 4  & 5   &                    0.7695 &     0.8727 & 0.8427 &        0.8630 &      0.8273  & 0.8350 \\
  128 & 4  & 2  & 12  &                    0.8076 &     0.8668 & 0.8616 &        0.8613 &      0.8409  & 0.8476 \\
  128 & 4  & 2  & 5   &                    0.7646 &     0.8658 & 0.8030 &        0.8516 &      0.8286  & 0.8227 \\
  128 & 4  & 4  & 12  &                    0.8143 &     0.8723 & 0.8414 &        0.8600 &      0.8639  & \textbf{0.8504} \\
  128 & 4  & 4  & 5   &                    0.7702 &     0.8640 & 0.8555 &        0.8575 &      0.8407  & 0.8376 \\
 \hline
\end{tabular}
 \caption{Hyperparameter tuning results, showing AUC ROC for each health condition, as well as the average across all health conditions. Here, Width is the number of output channels in each convolutional and LSTM layer, Conv Depth is the number of convolutional layers, LSTM Depth is the number of LSTM layers, and Initial Filter is the initial filter length of the first convolutional layer.}
\end{table*}

\section{Discussion}

We've shown that two methods, semi-supervised sequence learning and heuristic pretraining,
address a key technical challenge in applying deep learning to medicine: achieving high accuracy
with limited labeled data. Furthermore, we've shown high accuracy at detecting four common health
conditions---diabetes, sleep apnea, hypertension, and high cholesterol--using readily available,
off-the-shelf heart rate monitors. Since these four conditions are highly prevalent,
commonly undiagnosed, and reduced HRV is associated with the earliest stages of disease progression
\cite{schroeder2005diabetes,schroeder2003hypertension},
this suggests a new approach to public health screening using deep learning.

While these results are promising, artificial intelligence-based systems have shown high accuracy
at medical tasks---even outperforming human physicians---since the days of MYCIN \cite{yu1979antimicrobial},
CADUCEUS \cite{banks1985artificial}, and INTERNIST-I \cite{wolfram1995appraisal}. For AI-based systems to broadly improve human
health, several key challenges beyond prediction accuracy must be addressed.

A first challenge is confounding factors. Statistical models in medicine are typically used
for etiology, not just prediction, and cause (disease) and effect (treatment) may
be intertwined in subtle ways within medical data sets \cite{shmueli2010explain}. For example, a person with
previously-diagnosed hypertension may be prescribed beta blockers, which affect heart rate
variability, and therefore an algorithm with high prediction accuracy at identifying hypertension
may be picking up signals of both the underlying disease and the treatment. While there are
well-known algorithms to adjust for confounders in linear models, confounder-aware neural network
architectures are a critical future area of research.

A second challenge is deployment. While hospital departments such as the intensive care
unit or radiology department often have ample training data available, deploying a new algorithm
in a hospital is typically slow or impossible because errors are costly (a severe error may cost
a human life), the technical implementation is cumbersome (an integration with an electronic
medical record system), the organizational complexity is high (high regulatory burden, necessity to
train physicians on new systems), and the financial incentives are often misaligned (for
example, preventing readmissions may reduce the hospital's revenue). An alternative is to take
an outside-in approach: design care pathways for ambulatory, mobile-based screening that operate
\textit{outside} of the hospital but guide participants \textit{in} to the appropriate point of
care when a neural network detects high risk. Here, false negatives represent the status quo
(no diagnosis), and with sufficiently high precision the cost of false positives (e.g., blood tests,
mobile ECGs) is exceeded by the cost savings to the healthcare system of each true positive. Since
an algorithm can be packaged in a mobile app, it can be updated and
quickly deployed to participants. We are currently deploying our algorithm in ambulatory,
mobile trials in partnership with major medical centers to show real-world efficacy and build a
rigorous clinical evidence base \cite{tison2017A}.

A third challenge is interpretability. Deep learning systems perform well but are ``black boxes,''
whereas medical systems prefer to provide an explanation of the chain of reasoning. We intend to
incorporate mechanisms such as differentiable attention \cite{xu2015show,chan2016listen} to help both physicians
and regulators understand the internal workings of the neural network and its potential failure
modes.

A last challenge is the time-scale and volume of data. Many participants in our study have more than
a year of heart rate and step count data, corresponding to nearly one million time steps. We intend
to test neural architectures that can better model long-range dependencies, such as Clockwork RNNs
\cite{koutnik2014clockwork}, Phased LSTMs \cite{neil2016phased}, and Gaussian Process RNNs
\cite{futoma2017} in order to better model changes to a participant's physiological state over a long
period of time.

While these challenges may seem daunting, the constraints of medicine---scarce labeled data,
confounding factors, difficulty of deployment, and the necessity of interpretability---also
represent opportunities to develop new techniques. This work is a first step in showing how
health conditions can be detected using techniques first developed in natural language processing and computer vision.
Over time, we plan to extend this work to address the challenges above and deploy neural networks
that can improve health in the real world.

\bibliographystyle{aaai}

\bibliography{aaai_submission}
\end{document}